\pdfoutput=1

\documentclass[11pt]{article}

\usepackage[final]{EMNLP2023}

\usepackage{times}
\usepackage{latexsym}

\usepackage[T1]{fontenc}

\usepackage[utf8]{inputenc}

\usepackage{microtype}

\usepackage{inconsolata}
\usepackage{adjustbox} 
\usepackage{enumitem}
\usepackage{multirow}
\usepackage[utf8]{inputenc}
\usepackage{xcolor,colortbl}
\usepackage{microtype}
\usepackage{graphicx}
\usepackage{inconsolata}
\usepackage{booktabs, subcaption}
\title{ Machine Translation Advancements of Low-Resource Indian Languages by Transfer Learning}

\author{
  Bin Wei, Jiawei Zhen, Zongyao Li, Zhanglin Wu, Daimeng Wei, \\
  \textbf{Jiaxin Guo, Zhiqiang Rao,  Shaojun Li, Yuanchang Luo, Hengchao Shang,}   \\
  \textbf{
  Jinlong Yang, Yuhao Xie, Hao Yang
  }\\
  Huawei Translation Service Center, Beijing, China\\
  \{weibin29, zhengjiawei15, lizongyao, wuzhanglin2, weidaimeng, \\ guojiaxin1,raozhiqiang, lishaojun18, luoyuanchang1, shanghengchao,\\ 
  yangjinlong7, xieyuhao2, yanghao30\}@huawei.com 
  \\
  }

\begin{document}
\maketitle
\begin{abstract}

This paper introduces the submission by Huawei Translation Center (HW-TSC) to the WMT24 Indian Languages Machine Translation (MT) Shared Task. To develop a reliable machine translation system for low-resource Indian languages, we employed two distinct knowledge transfer strategies, taking into account the characteristics of the language scripts and the support available from existing open-source models for Indian languages. For Assamese(as) and Manipuri(mn), we fine-tuned the existing IndicTrans2\cite{gala2023indictrans2} open-source model to enable bidirectional translation between English and these languages. For Khasi (kh) and Mizo (mz), We trained a multilingual model as a baseline using bilingual data from these four language pairs, along with an additional about 8kw English-Bengali bilingual data, all of which share certain linguistic features. This was followed by fine-tuning to achieve bidirectional translation between English and Khasi, as well as English and Mizo. Our transfer learning experiments produced impressive results: 23.5 BLEU for en→as, 31.8 BLEU for en→mn, 36.2 BLEU for as→en, and 47.9 BLEU for mn→en on their respective test sets. Similarly, the multilingual model transfer learning experiments yielded impressive outcomes, achieving 19.7 BLEU for en→kh, 32.8 BLEU for en→mz, 16.1 BLEU for kh→en, and 33.9 BLEU for mz→en on their respective test sets. These results not only highlight the effectiveness of transfer learning techniques for low-resource languages but also contribute to advancing machine translation capabilities for low-resource Indian languages.

\end{abstract}

\section{Introduction}

In the realm of machine translation, Neural Machine Translation (NMT) has become the dominant technology, as confirmed by previous research. However, training NMT models requires large amounts of data, which presents a significant challenge when dealing with low-resource languages. To tackle this challenge, we employed transfer learning, a well-established approach that enhances model performance by transferring knowledge gained from one task to other related tasks. To improve translation capabilities for low-resource languages, we faced the challenge of limited bilingual resources for Indian languages. To overcome this issue, we trained a multilingual model using not only all the bilingual data provided for the task but also additional Bengali data. Additionally, we examined the languages supported by the existing IndicTrans2\cite{gala2023indictrans2} open-source model and conducted a comparative analysis. Based on our findings, we selected different baseline models for knowledge transfer depending on the language pair: for Assamese and Manipuri, we used the IndicTrans2 model as the baseline, while for Khasi and Mizo, we trained multilingual model by ourselves as the baseline. This approach enabled us to effectively leverage existing resources while addressing the specific challenges associated with each language pair.



IndicTrans2 is the first open-source transformer-based multilingual NMT model that supports high-quality translations across all the 22 scheduled Indic languages. It was trained on the extensive Bharat Parallel Corpus Collection (BPCC), a publicly accessible repository encompassing both pre-existing and freshly curated data for all 22 scheduled Indian languages, this model boasts a comprehensive understanding of the linguistic diversity within the Indian subcontinent. To enhance its linguistic prowess, IndicTrans2 has undergone auxiliary training utilizing the rich resource of back-translated monolingual data. The model was then trained on human-annotated data to achieve further improvements. We used this model in the first two subtasks and fine-tuned it on the training data provided by WMT24. By adopting this approach, we aim to capitalize on the acquired knowledge during training to significantly bolster the performance of the model in the specific translation task at hand. The fine-tuned IndicTrans2 achieves good scores, so we are using it for our final submission in the first two subtasks. 




For the multilingual model, we first utilized resources from Bengali. The choice of Bengali was based on its belonging to the Indo-Aryan branch, its linguistic feature similarities with some of the target low-resource languages, and its relatively rich available data. By introducing Bengali data, we aimed to enable the model to learn features potentially shared with the target languages, thereby laying a foundation for processing other related languages. Next, we integrated all available bilingual data from Indic language MT track. This included parallel corpora between various Indian languages and English. Although the data for each language pair might be limited individually, the combined dataset offered diverse learning samples. We believe that this integration of multilingual data helps the model capture both the commonalities and differences among different Indian languages. Based on this carefully selected and integrated data, we trained a multilingual model. The design goal of this model was to handle translation tasks for multiple Indian languages simultaneously, using the commonalities between languages to compensate for the scarcity of data in any single language. Through this approach, we expect the model to learn more generalized language representations and translation knowledge under resource constraints, leading to improved performance on Khasi and Mizo translation tasks.

Ultimately, we adopted a differentiated strategy for knowledge transfer. This approach thoroughly considered the characteristics of each language to achieve optimal transfer effects. In Section 2, we will discuss the details of the data, the methods and processes used for data pre-processing. Section 3 will cover the overall architecture and training strategies of the NMT system, including a detailed account of the various optimization methods. In Section 4, we will present the experimental parameters, results, and their analysis. The final section will summarize the key findings of the paper.

\section{Data}

\subsection{Data Details} 


We have fine-tuned the model using the WMT24 corpus. Additionally, we used 2M monolingual english dataset to do BT and FT. The amount of data we used is shown in Tables \ref{data}.

\begin{table}[ht]
\begin{center}
\begin{adjustbox}{width=\columnwidth,center}
\begin{tabular}{@{}ccc@{}}
\toprule  
language pairs &  bitext data & monolingual data \\
\hline
en-as & 50K & en: 2M, as: 2.62M\\
\hline
en-mn & 21K & en: 2M, mn: 2.14M\\
\hline
en-kh & 24K & en: 2M, kh: 182K\\
\hline
en-mz & 50K & en: 2M, mz: 1.9M\\
\bottomrule 
\end{tabular}
\end{adjustbox}
\caption{Bilingual and monolingual used for training NMT models.}\label{data}
\end{center}
\end{table}

\subsection{Data Pre-processing}

Our data pre-processing methods for NMT include: 
\begin{itemize}
    \item Remove duplicate sentences or sentence pairs.
    \item Convert full-width symbols to half-width.
    \item Use fasttext\footnote{\url{https://github.com/facebookresearch/fastText}} \cite{joulin2016fasttext} to filter other language sentences.
    \item Use mosesdecoder\footnote{\url{https://github.com/moses-smt/mosesdecoder}} \cite{koehn-etal-2007-moses} to normalize English punctuation.
    \item Filter out sentences with more than 150 words.
    \item Use fast-align \cite{dyer2013simple} to filter sentence pairs with poor alignment.
    \item Sentencepiece\footnote{\url{https://github.com/google/sentencepiece}} (SPM) \cite{kudo2018sentencepiece} is used to perform subword segmentation, and the vocabulary size is set to 32K.
\end{itemize}

Since there may be some semantically dissimilar sentence pairs in bilingual data, we use LaBSE\footnote{\url{https://huggingface.co/sentence-transformers/LaBSE}} \cite{feng2022language} to calculate the semantic similarity of each bilingual sentence pair, and exclude bilingual sentence pairs with a similarity score lower than 0.75 from our training corpus.

\begin{figure}[t] 
\centering
\includegraphics[width=90mm,height=60mm]{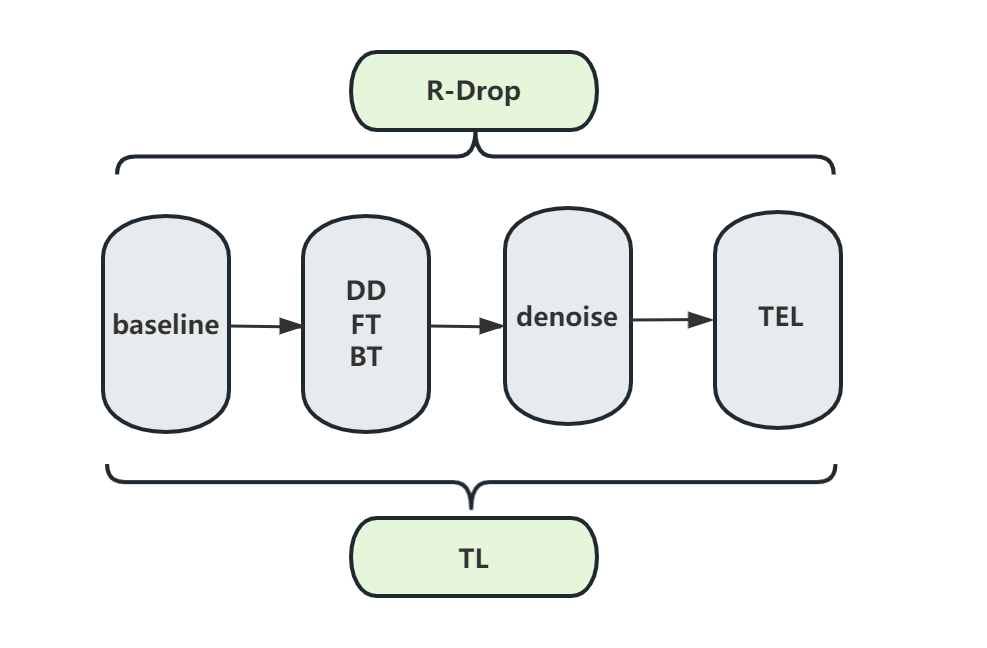}
\caption{\centering The overall training flow of NMT system.}
\label{NMT_TRAIN}
\end{figure}

\section{NMT System}

\subsection{System Overview}

We use Transformer \cite{vaswani2017attention} as our neural machine translation (NMT)\cite{Bahdanau2014NeuralMT} model architecture. For the first two subtasks(en-as, en-mn), we use the IndicTrans2\cite{gala2023indictrans2} model as our baseline model, which is a deep Transformer architecture with 18-layers encoder and 18-layers decoder. with the latter two subtasks(en-kh, en-mz), we trained a multilingual model as our baseline model, which is a deep Transformer architecture with 35-layers encoder and 3-layers decoder.

Fig. \ref{NMT_TRAIN} shows the overall training flow of NMT system. Referred to previous work \cite{wei2021hw,wei2022hw,wu2023path}, We use training strategies such as regularized dropout (R-Drop) \cite{wu2021r}, data diversification (DD) \cite{nguyen2020data}, forward translation {FT) \cite{abdulmumin2021enhanced}, back translation (BT) \cite{sennrich2016improving}, denoise, Transfer learning(TL) and transductive ensemble learning (TEL) \cite{wang2020transductive} for training.


\subsection{Regularized Dropout}

Regularized Dropout (R-Drop)\footnote{\url{https://github.com/dropreg/R-Drop}} \cite{wu2021r} is a simple yet more effective alternative to regularize the training inconsistency induced by dropout \cite{srivastava2014dropout}. Concretely, in each mini-batch training, each data sample goes through the forward pass twice, and each pass is processed by a different sub model by randomly dropping out some hidden units. R-Drop forces the two distributions for the same data sample outputted by the two sub models to be consistent with each other, through minimizing the bidirectional Kullback-Leibler (KL) divergence \cite{van2014renyi} between the two distributions. That is, R-Drop regularizes the outputs of two sub models randomly sampled from dropout for each data sample in training. In this way, the inconsistency between the training and inference stage can be alleviated.

\subsection{Data Diversification}

Data Diversification (DD) \citep{nguyen2020data} is a data augmentation method to boost NMT performance. It diversifies the training data by using the predictions of multiple forward and backward models and then merging them with the original dataset which the final NMT model is trained on. DD is applicable to all NMT models. It does not require extra monolingual data, nor does it add more parameters. To conserve training resources, we only use one forward model and one backward model to diversify the training data.

\subsection{Forward Translation}

Forward translation (FT) \cite{abdulmumin2021enhanced}, also known as self-training, is one of the most commonly used data augmentation methods. FT has proven effective for improving NMT performance by augmenting model training with synthetic parallel data. Generally, FT is performed in three steps: (1) randomly sample a subset from the large-scale source monolingual data; (2) use a “teacher” NMT model to translate the subset data into the target language to construct the synthetic parallel data; (3) combine the synthetic and authentic parallel data to train a “student” NMT model.

\subsection{Back Translation}

An effective method to improve NMT with target monolingual data is to augment the parallel training data with back translation (BT) \cite{sennrich2016improving,wei-etal-2023-text}. There are many published works that expand the understanding of BT and investigate methods for generating synthetic source sentences. \citeauthor{edunov2018understanding} (\citeyear{edunov2018understanding}) find that back translations obtained via sampling or noised beam outputs are more effective than back translations generated by beam or greedy search in most scenarios. \citeauthor{caswell2019tagged} (\citeyear{caswell2019tagged}) show that the main role of such noised beam outputs is not to diversify the source side, but simply to tell the model that the given source is synthetic. Therefore, they propose a simpler alternative strategy: Tagged BT. This method uses an extra token to mark back translated source sentences, which generally outperforms noised BT \cite{edunov2018understanding}. For better joint use with FT, we use sampling back translation (ST) \cite{edunov2018understanding}.

\subsection{Denoise} 

In machine translation, denoising improves translation quality by removing noise from the training data, such as inaccurate translations, grammatical errors, or unnatural sentence structures, allowing the model to focus on high-quality data and produce more accurate and fluent translations. Additionally, denoising enhances the model's robustness by eliminating noisy data, which helps the model better learn the target language's patterns, reducing errors and leading to more stable and reliable performance across diverse inputs. It also optimizes training efficiency by decreasing the amount of data the model needs to process, particularly by filtering out low-quality data, which results in a cleaner and more consistent dataset and can shorten the overall training time. Moreover, denoising reduces error propagation by preventing the model from learning incorrect language patterns, thereby minimizing the accumulation and spread of errors in generated translations. Finally, it enhances the model's generalization ability, as denoised data is more representative, enabling the model to better adapt to different types of input sentences and improving its performance in real-world applications. Through denoising, machine translation models can more effectively utilize high-quality data, leading to superior translation outcomes and greater overall model stability.

\subsection{Transductive Ensemble Learning}

Ensemble learning \cite{garmash2016ensemble}, which aggregates multiple diverse models for inference, is a common practice to improve the performance of machine learning models. However, it has been observed that the conventional ensemble methods only bring marginal improvement for NMT when individual models are strong or there are a large number of individual models. Transductive Ensemble Learning (TEL) \cite{zhang2019curriculum} studies how to effectively aggregate multiple NMT models under the transductive setting where the source sentences of the test set are known. TEL uses all individual models to translate the source test set into the target language space and then finetune a strong model on the translated synthetic data, which significantly boosts strong individual models and benefits a lot from more individual models.

\subsection{Transfer Learning}
Transfer learning(TL) is a machine learning technique where a model trained on one task is adapted for a second related task. Instead of starting the training of a new model from scratch, transfer learning leverages the knowledge learned from the first task to improve learning on the second task. For Assamese(as) and Manipuri(mn), We have used IndicTrans2\cite{gala2023indictrans2}, a powerful model that performs well for English-to-Indic and Indic-to English translation for 22 scheduled Indian languages. This knowledge can be used to translate other Indian languages to and from English. Our approach entailed the fine-tuning of this model, leveraging the parallel corpus provided by the WMT24 for the Indic MT task. This fine-tuning process equipped the model with the expertise required to proficiently translate Assamese and Manipuri to and from English, ultimately yielding the most outstanding results. Similarly, for Khasi and Mizo, we trained a multilingual model as the baseline. We also applied transfer learning techniques to enhance the baseline model using data specific to these language pairs. The results on both the test and dev sets were highly encouraging.

\section{Experiment}

\subsection{Settings}

We use Transformer architecture in all the subtasks. For the first two subtasks, we use IndicTrans2 \cite{gala2023indictrans2} as our baseline model, which is a deep Transformer architecture with 18-layers encoder and 18-layers decoder. With the latter subtasks, the model is also a Transformer architecture with 35-layers encoder and 3-layers decoder. For the first two subtasks, our models apply Adam \cite{kingma2014adam} as optimizer to update the parameters with $\beta_1$ = 0.9 and $\beta_2$ = 0.98. We employ a warm-up learning rate of $10^{-7}$ for 2000 update steps and a learning rate of $3 \ast10^{-5}$. For normalization, we use a dropout value of 0.2 and normalize the probabilities using smoothed label cross-entropy. We use GeLU activations \cite{hendrycks2016bridging} for better learning. For the latter subtasks, parameter update frequency is 2, and learning rate is 5e-4. The number of warmup steps is 4000, and model is saved every 1000 steps. R-Drop \cite{wu2021r} is used in model training for all subtasks, and we set $\lambda$ to 5.

We use the scareBLEU library to calculate our BLEU \cite{papineni2002bleu} and ChrF \cite{popovic2015chrf} scores with a word order of 2.

\subsection{Results}
Regarding this four language pair directions, we use Regularized Dropout, Bidirectional Training, Data Diversification, Forward Translation, Back Translation, Alternated Training, Curriculum Learning, and Transductive Ensemble Learning. The evaluation results of four language pair directions NMT system on WMT24 Indic MT test and dev set are shown in Tables \ref{result1} and Tables \ref{result2}. 

As shown in Table \ref{result1}, IndicTrans2\cite{gala2023indictrans2} provides a strong baseline. Fine-tuning the model with FT, BT, and bitext data leads to significant improvements, particularly in the en-mn direction, where the BLEU score increases by nearly 20 points over the baseline on the test and dev set. This improvement is largely attributed to Data Diversification. Table \ref{result2} further illustrates that FT and BT data contribute the most to model performance, especially in the en-mz direction, which sees an increase of nearly six BLEU points compared to the multilingual baseline. Even after enhancing the model with BT and FT data, adding filtered high-quality bilingual data results in an average gain of about one BLEU point, highlighting the critical role of data quality. Finally, we all use TEL technique to obtain a good result, the improvement is very small, almost less than one bleu score.

\begin{table}
\begin{center}
\begin{adjustbox}{width=\columnwidth,center}
\begin{tabular}{@{}clcccccc@{}}
\toprule  
 Language-pair & Training strategies & Bleu(test) & ChrF2(test) & Bleu(dev) & ChrF2(dev)\\
\hline
\multirow{4}*{en$\rightarrow$as} & IndicTrans2 baseline & 18.9 & 51.4 & 14.7 & 44.8\\
{ } & + DD, FT, BT & 22.9 & 52.5& 21.1 & 47.7\\
{ } & + denoise & 23.3 & 53.1 & 22.5 & 48.9\\
{ } & + TEL & 23.5 & 53.2 & 22.8 & 49.0\\
\hline
\multirow{4}*{en$\rightarrow$mn} & IndicTrans2 baseline & 11.9 & 48.5 & 11.9 & 48.5\\
{ } & + DD, FT, BT & 30.9 & 62.8 & 31.1 & 63.4\\
{ } & + denoise & 31.7 & 64.7 & 31.7 & 64.9\\
{ } & + TEL & 31.8 & 64.6 & 31.6 & 64.9\\
\hline
\multirow{4}*{as$\rightarrow$en} & IndicTrans2 baseline & 29.7 & 56.3 & 25.6 & 49.3\\
{ } & + DD, FT, BT & 35.8 & 58.6 & 35.0 & 54.5\\
{ } & + denoise & 36.1 & 58.6 & 34.8 & 54.6\\
{ } & + TEL & 36.2 & 59.4 & 33.7 & 54.2\\
\hline
\multirow{4}*{mn$\rightarrow$en} & IndicTrans2 baseline & 32.6 & 62.3 & 33.4 & 61.8\\
{ } & + DD, FT, BT & 47.5 & 70.8 & 47.0 & 69.7\\
{ } & + denoise & 47.7 & 70.8 & 47.2 & 69.7\\
{ } & + TEL & 47.9 & 70.8 & 47.4 & 69.8\\
\bottomrule 
\end{tabular}
\end{adjustbox}
\caption{The results of en-as and en-mn language pairs on the test and dev set.}\label{result1}
\end{center}
\end{table}

\begin{table}
\begin{center}
\begin{adjustbox}{width=\columnwidth,center} 
\begin{tabular}{@{}clcccccc@{}}
\toprule  
 Language-pair & Training strategies & Bleu(test) & ChrF2(test) & Bleu(dev) & ChrF2(dev) \\
\hline
\multirow{4}*{en$\rightarrow$kh} & multilingual baseline & 17.4 & 40.4 & 17.0 & 39.7\\
{ } & + DD, FT, BT & 18.1 & 41.8 & 17.9 & 41.3\\
{ } & + denoise & 19.5 & 43.3 & 19.2 & 42.7\\
{ } & + TEL & 19.7 & 43.5 & 19.3 & 42.8\\
\hline
\multirow{4}*{en$\rightarrow$mz} & multilingual baseline & 25.0 & 51.6 & 22.3 & 46.6\\
{ } & + DD, FT, BT & 30.8 & 55.7 & 25.2 & 49.1\\
{ } & + denoise & 32.5 & 57.1 & 25.4 & 49.3\\
{ } & + TEL & 32.8 & 57.3 & 25.7 & 49.4\\
\hline
\multirow{4}*{kh$\rightarrow$en} & multilingual baseline & 15.1 & 37.7 & 15.0 & 38.1\\
{ } & + DD, FT, BT & 15.8 & 37.8 & 15.0 & 38.3\\
{ } & + denoise & 15.9 & 38.5 & 15.5 & 39.0\\
{ } & + TEL & 16.1 & 38.8 & 15.6 & 39.2\\
\hline
\multirow{4}*{mz$\rightarrow$en} & multilingual baseline & 26.7 & 48.2 & 22.9 & 44.0\\
{ } & + DD, FT, BT & 32.9 & 52.2 & 25.0 & 45.4\\
{ } & + denoise & 33.7 & 52.2 & 25.8 & 46.5\\
{ } & + TEL & 33.9 & 52.7 & 26.0 & 46.7\\
\bottomrule 

\end{tabular}
\end{adjustbox}
\caption{The results of en-kh and en-mz language pairs on the test and dev set.}\label{result2}
\end{center}
\end{table}

\section{Conclusion}

This paper presents the submission of HW-TSC to the WMT24 Indic MT Task. For the first two subtasks, we use IndicTrans2 as our baseline model to fine-tune it with corpus provided by WMT24 on the en-as and en-mn language pairs, which achieves remarkable performance. For the latter two subtasks, we train a multilingual model on the en-kh and en-mz language pairs, and then use training strategies such as R-Drop, DD, FT, BT, denoise and TEL to train the NMT model based on the deep Transformer-big architecture. By applying these training strategies, our submission achieved a competitive result in the final evaluation.  

\bibliography{emnlp2023}
\bibliographystyle{acl_natbib}

\end{document}